\documentclass{article}
\usepackage{ijcai26}
\usepackage{times} %
\usepackage{soul}
\usepackage{url}
\usepackage[hidelinks]{hyperref}
\usepackage[utf8]{inputenc}
\usepackage[small]{caption}
\usepackage{graphicx}
\usepackage{amsmath}
\usepackage{amsthm}
\usepackage{booktabs}
\usepackage[switch]{lineno}
\usepackage{epigraph}
\usepackage{color}

\title{Low-Latency Real-Time Audio Game Commentary System\\via LLM-Based Parallel Text Generation}

\author{
Ryota Kawamatsu$^{1,2}$
\and
Anum Afzal$^{2,3}$\and
Yuki Saito$^1$\and
Shinnosuke Takamichi$^{4,1}$\and \\
Graham Neubig$^5$\and
Katsuhito Sudoh$^6$\and
Hiroya Takamura$^2$\and
Tatsuya Ishigaki$^{2}$\\
\affiliations
$^1$The University of Tokyo, Japan\\
$^2$National Institute of Advanced Industrial Science and Technology, Japan\\
$^3$Technical University of Munich, Germany\\
$^4$Keio University, Japan\\
$^5$Carnegie Mellon University, U.S.A.\\
$^6$Nara Women's University, Japan
}

\begin{document}

\maketitle

\begin{abstract}
We present a low-latency real-time audio game commentary system that generates spoken commentary directly from live gameplay video.
In this end-to-end setting, a key bottleneck is accumulated waiting time; conventional pipelines capture frames, generate text, and synthesize speech sequentially for each utterance, and do not request the next generation until speech playback has completed.
This strict sequentiality causes long and unnatural silence between utterances.
To address this latency bottleneck, our system runs text generation in parallel with speech playback and buffers multiple candidate utterances ahead of time, enabling immediate synthesis at playback boundaries.
Experiments on fast-paced game videos show that our parallel design reduces the mean inter-utterance silence from 9.6 seconds to 0.3 seconds compared to sequential baselines.
It also improves similarity to professional speaking--silence timing patterns by over 40~\%, and a user study with 120 experienced game players confirms significantly improved perceived speaking rhythm.
Our demo video is available at: \textcolor{blue}{\url{https://youtu.be/pmrRUlvav8M}}.
\end{abstract}

\begin{figure}[t]
    \centering
    \includegraphics[width=\linewidth]{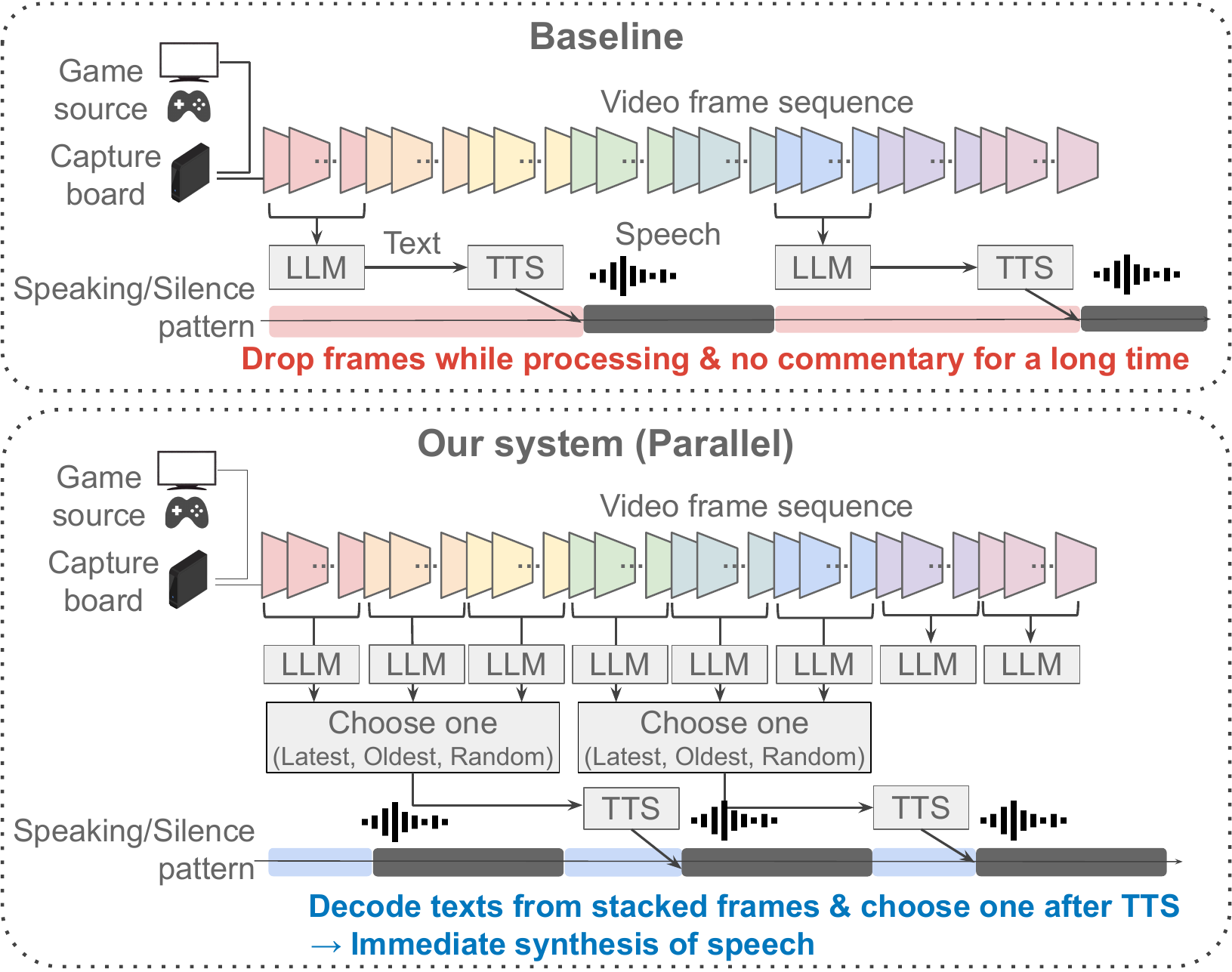}
    \caption{System overview: sequential baseline vs. our parallel generation with buffering and video delay control.}
    \label{fig:system}
\end{figure}

\section{Introduction}

Live game commentary describes in-game events and provides contextual explanations to help viewers immerse themselves in gameplay~\cite{Behrens02012022,AVisionfortheFormalDocumentationandDigitalizationofSportsCommentatorsCommentaries}. Audio commentary is integrated with the live-streamed video and delivered to users via online streaming platforms.
Advances in multimodal language models and audio synthesis have accelerated research on automatic audio commentary generation~\cite{zheng2025multimodalperceptionstrategicreasoning}.
Prior work has explored language generation from structured data~\cite{taniguchi-etal-2019-generating,ishigaki-etal-2021-generating}, video~\cite{yu_sports_narrative,Kim_2020_baseball,rao-etal-2024-matchtime,Zhou_2024_CVPR,Chen_2025_CVPR,afzal2026realtimegenerationgamevideo}, and multimodal inputs~\cite{mori-etal-2025-live,someya-etal-2025-live,wang-yoshinaga-2025-commentary},
as well as expressive speech synthesis for engaging commentary\cite{Iura_access};
however, these two lines of research have mostly progressed independently.

Building on these separate research directions,
several works have begun to combine language generation and speech synthesis
into end-to-end audio commentary systems~\cite{kumano_2019_e2e,Jumneanbun_2020_CoG,Xu_2021_HighlightTTS,ishigaki-etal-2023-audio}. However, existing systems rely on structured game state inputs and lightweight language models, which prevent latency from becoming a critical bottleneck.
Extending this setting to raw gameplay video and modern multimodal large language models (LLMs) introduces substantial latency challenges~\cite{mori-etal-2025-live}.

A major bottleneck in real-time audio commentary is its pipeline design.
Most pipelines are sequential: 
they process game inputs, generate text, synthesize speech, and start generating the next utterance only after the current speech playback has completed. As a result, waiting time accumulates across turns and leads to long silence between utterances, which substantially degrades perceived commentary quality.

To address this latency bottleneck, we propose an audio commentary system with parallel text generation and a lightweight control of video delay.
While one utterance is being spoken, our system continues generating candidate utterances in the background process for newly arriving video segments and stores them in a buffer. When the current speech playback ends, the system immediately selects a buffered candidate and synthesizes speech from it, thereby avoiding silence caused by waiting for generation to finish. Since inference latency is inevitable in practice, we additionally implement a lightweight video delay control that intentionally delays the outgoing video stream to better align generated speech with the displayed visuals.

Experiments on fast-paced game videos~\cite{saito-etal-2020-smash} show that our system reduces mean inter-utterance silence from 9.6 seconds to 0.3 seconds, 
a gain further supported by a human evaluation with 120 participants.

\section{Real-Time Game Audio Commentary}
A typical real-time audio commentary pipeline converts a gameplay stream into audio commentary by executing (i) video acquisition and segmentation, (ii) text generation, and (iii) speech synthesis.
The video stream is buffered and grouped into short frame segments. Let $f_i$ denote the frame captured at time $i$; then a segment is $F_k = \{f_i,\ldots,f_{i+N-1}\}$, which consists of $N$ consecutive frames.
A common design is to run these modules sequentially. Specifically, the system waits until the current speech playback ends, and then calls a multimodal LLM on the latest segment to generate text using a simple prompt (e.g., ``Describe the game situation. If you have nothing to say, stay silent.'')~\cite{afzal2026realtimegenerationgamevideo}.
The generated text is then converted into audio by a text-to-speech (TTS) module~\cite{Iura_access}.
Because the next generation is triggered only after playback finishes, this strict sequentiality accumulates latency and causes long silences between utterances.

\section{Proposed System}
We propose a low-latency system design based on two key ideas: parallel text generation (our main mechanism to reduce waiting time) and lightweight video delay control (to maintain temporal consistency).

\subsection{Parallel Text Generation}
Unlike conventional systems that trigger text generation only after speech playback ends, our system initiates text generation as soon as a new video segment becomes available, even while the current utterance is still being spoken (Figure~\ref{fig:system}).
As a result, multiple candidate utterances are generated ahead of playback and stored in a buffer.
When the current utterance ends, the system selects one buffered candidate and synthesizes it immediately,
eliminating idle silence caused by sequential waiting.

We consider three lightweight selection policies: \emph{Latest} (most recent segment), \emph{Oldest} (earliest buffered), and \emph{Random}.
We adopt these lightweight policies to avoid additional decision latency in real-time settings.

\subsection{Video Delay Control}
\label{sec:delay_control}

We consider a streaming setup for live streaming platforms such as YouTube Live, where gameplay video and automatically generated speech are integrated into a single audiovisual stream on our server and then uploaded to the platform. This setting allows us to control video playback timing to mitigate misalignment caused by generation latency.

Since text generation and TTS inevitably introduce delays that cause commentary to lag, we absorb this latency by intentionally delaying the video. Specifically, we buffer the video stream and start playback only when the first utterance begins, matching the initial end-to-end generation latency.

\section{Experiments}

\noindent
\textbf{Demo Setups: }In the demonstration, conference attendees voluntarily operate a Nintendo Switch console on-site, and the gameplay video is captured in real time using a capture device (Elgato HD60 X\footnote{\url{https://www.elgato.com/jp/ja/p/game-capture-hd60-x}}) and streamed to a commentary server.
The system generates spoken commentary online based on the live video input.
The demo is conducted exclusively on-site, and no gameplay video or audio is recorded or distributed.
Attendees observe the gameplay together with automatically generated commentary, allowing them to directly perceive how parallel candidate generation affects inter-utterance silence and speaking rhythm.
The gameplay machine sends video at 25 fps, and captured frames are grouped into segments of $N$ consecutive frames.
In all experiments, we set $N=32$.
Each segment $F_k = \{f_i, \ldots, f_{i+N-1}\}$ is sent to the commentary server as soon as it becomes available.
The language model receives each segment as base64-encoded tokens and generates commentary text.
We use \textit{GPT-4.1-mini}
\footnote{https://platform.openai.com/docs/models/gpt-4.1-mini} %
for text generation and follow the TTS configuration described in~\cite{Iura_access}.
Our experiments vary the \texttt{max\_new\_tokens} parameter in $\{20, 40, 60, 80, 100\}$.

\noindent
\textbf{Dataset: }We randomly selected 8 videos from the Smash Corpus~\cite{saito-etal-2020-smash} whose target game is \textit{Super Smash Bros.\ Ultimate}.
We select this game for our experiments because it is a fast-paced game where latency is particularly noticeable. This corpus provides gameplay video. We collected commentaries from skilled commentators for the selected videos, enabling a comparison between skilled human- and system-generated commentaries.

\noindent
\textbf{Compared Models: }We compare five methods.
All methods use the same video delay control (Section~\ref{sec:delay_control}).
\emph{Baseline After-Audio} is a fully sequential pipeline that starts generating the next utterance only after the current speech playback ends, following a common design in prior real-time commentary systems~\cite{ishigaki-etal-2023-audio}.
\emph{Baseline After-Text} is a semi-sequential variant that starts generating the next utterance immediately after text generation finishes, without waiting for speech playback to end.
Our system uses parallel generation with buffering, and we test three lightweight selection policies: \emph{Parallel Latest}, \emph{Prallel Oldest}, and \emph{Parallel Random}.

\begin{table}[t]
\centering
\small
\setlength{\tabcolsep}{3pt}
\begin{tabular}{lccc}
\hline
Method & Cumulative silence & Mean silence & mIoU \\
\hline
Human &
$59.7 \pm 11.3$ &
$1.7 \pm 0.4$ &
-- \\
\hline
After-Audio &
$134.0 \pm 5.2$ &
$9.5 \pm 0.9$ &
$0.01 \pm 0.06$ \\
After-Text &
$125.5 \pm 5.8$ &
$6.8 \pm 0.9$ &
$0.10 \pm 0.08$ \\
\hline
Latest &
$24.7 \pm 6.6$ &
$0.4 \pm 0.1$ &
$0.59 \pm 0.04$ \\
Oldest &
$\mathbf{18.9} \pm 3.1$ &
$\mathbf{0.3} \pm 0.1$ &
$\mathbf{0.60} \pm 0.04$ \\
Random &
$19.1 \pm 3.6$ &
$\mathbf{0.3} \pm 0.1$ &
$\mathbf{0.60} \pm 0.03$ \\
\hline
\end{tabular}
\caption{Silence statistics and mIoU for \texttt{max\_new\_tokens}=20. Values in parentheses denote standard deviation.}
\label{tab:timing_stats}
\end{table}

\begin{table}[t]
\centering
\small
\setlength{\tabcolsep}{3pt}
\begin{tabular}{lcc}
\hline
\texttt{max\_new\_tokens} & Mean speaking time & Text length \\
\hline
Human  & $2.8 \pm 0.5$ & $25.9 \pm 4.1$ \\
\hline
$20$   & $2.3 \pm 0.1$ & $16.8 \pm 0.4$ \\
$40$   & $2.9 \pm 0.1$ & $22.7 \pm 0.6$ \\
$60$   & $3.4 \pm 0.1$ & $28.2 \pm 0.8$ \\
$80$   & $3.8 \pm 0.1$ & $31.8 \pm 1.1$ \\
$100$  & $4.4 \pm 0.1$ & $36.3 \pm 1.3$ \\
\hline
\end{tabular}
\caption{Effect of \texttt{max\_new\_tokens} on speaking duration and text length (Latest policy).}
\label{tab:max_new_tokens}
\end{table}

\begin{figure}[t]
    \centering
    \includegraphics[width=\linewidth]{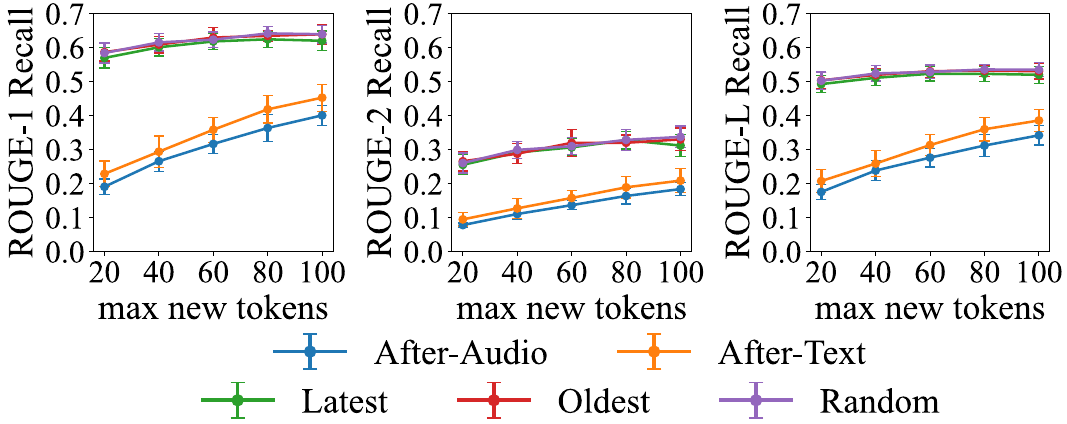}
    \caption{ROUGE Recall across \texttt{max\_new\_tokens}. Error bars denote standard deviation.}
\label{fig:different_max_tokens}
\end{figure}

\begin{figure}[t]
    \centering
    \includegraphics[width=0.9\linewidth]{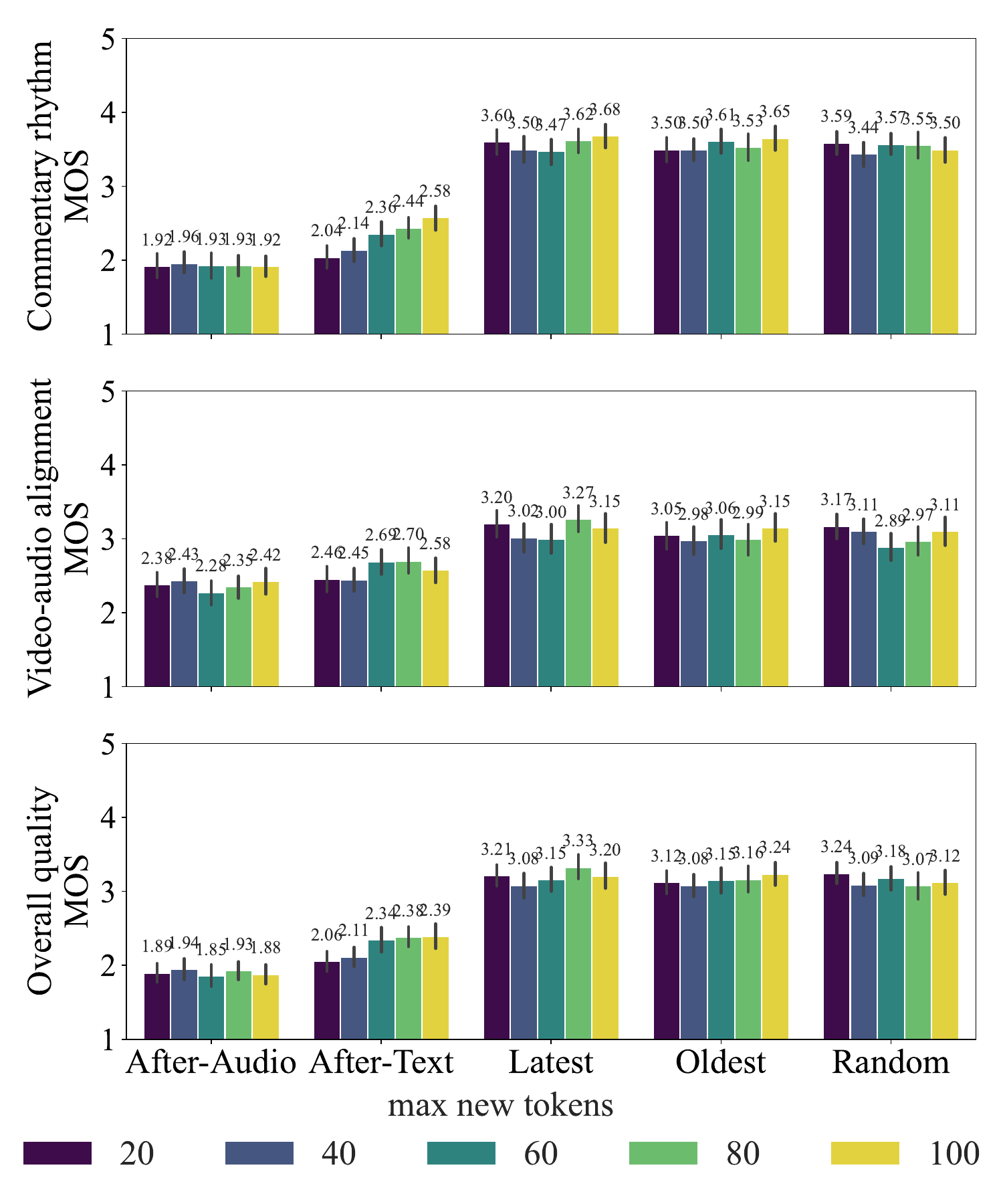}
    \caption{User study results for Q1--Q3 across methods and \texttt{max\_new\_tokens}. Error bars denote 95\% confidence intervals.}
    \label{fig:subj_eval}
\end{figure}

\noindent
\textbf{Evaluation Metrics: }We evaluate commentaries in terms of timing behavior and content adequacy.
For timing, we compare cumulative silence, mean silence between utterances, and utterance length.
To capture similarity of speaking and silence patterns to human-spoken commentaries, we represent each commentary stream as a 1\,Hz binary sequence
(1 = speaking, 0 = silence) and compute the mean Intersection over Union (mIoU) between automatically generated commentaries and skilled human-spoken ones.
Higher mIoU indicates fewer mismatches in speaking and silence regions compared to skilled human-spoken commentaries.
For content adequacy, we compute ROUGE~\cite{lin-2004-rouge} scores against reference commentary text.
Because exact temporal alignment is difficult in real-time settings, ROUGE is computed within fixed 10-second windows.

For human evaluation, we conduct a user study with 120 crowdworkers recruited via Lancers\footnote{\url{https://www.lancers.jp/}}. We present 30-second clips sampled from 20 scenes. Each clip is paired with commentary generated by one of 25 conditions (5 methods $\times$ 5 \texttt{max\_new\_tokens} settings). Participants rate on 5-point Likert scales, and we report Mean Opinion Scores (MOS) for the following criteria:
(Q1) commentary rhythm naturalness, (Q2) alignment with the video, and (Q3) overall quality.

\section{Results}

Table~\ref{tab:timing_stats} summarizes silence-related statistics and mIoU for each method.
The sequential baselines (After-Audio and After-Text) exhibit approximately twice the cumulative silence of human commentary
and extremely low mIoU scores, indicating that sequential processing introduces long and unnatural pauses between utterances.
In contrast, all proposed parallel methods substantially reduce silence, achieving mIoU values close to human commentary.
Differences among the selection policies are relatively small, suggesting that the main benefit comes from parallel generation itself.

Table~\ref{tab:max_new_tokens} analyzes the effect of \texttt{max\_new\_tokens} under the Latest policy.
While a setting of 40 best matches human speaking duration, 60 yields utterance lengths closest to human commentary,
indicating that generation length affects different aspects of temporal behavior.
ROUGE scores (Figure~\ref{fig:different_max_tokens}) show consistent improvements of parallel methods over sequential baselines.

Figure~\ref{fig:subj_eval} reports the results of the human evaluation.
The proposed methods significantly outperform the baselines across all criteria (Q1--Q3),
demonstrating that reducing unnatural silence leads to higher perceived commentary quality.

\section{Conclusion}
We presented a real-time game audio commentary system that reduces latency by parallel generation.
Experiments show substantial reductions in inter-utterance silence
and improved similarity to professional speaking/silence patterns.

\section*{Acknowledgements}
This paper is based on results obtained from a project, Programs for Bridging the gap between R\&D and the IDeal society (society 5.0) and Generating Economic and social value (BRIDGE)/Practical Global Research in the AI × Robotics Services, implemented by the Cabinet Office, Government of Japan.
This paper is also based on results obtained from AIST policy-based budget project ``R\&D on Generative AI Foundation Models for the Physical Domain''.

\bibliographystyle{named}
\bibliography{ijcai26}

\end{document}